# ML-Driven Approaches to Combat Medicare Fraud: Advances in Class Imbalance Solutions, Feature Engineering, Adaptive Learning, and Business Impact

Dorsa Farahmandazad[1], Kasra Danesh[2]

[1] *College of Business, Florida Atlantic University, Boca Raton, FL, USA*
[2]*Department of Electrical Engineering and Computer Science, Florida Atlantic University, Boca Raton, FL, USA*

## Abstract

Medicare fraud poses a substantial challenge to healthcare systems, resulting in significant financial losses and undermining the quality of care provided to legitimate beneficiaries. This study investigates the use of machine learning (ML) to enhance Medicare fraud detection, addressing key challenges such as class imbalance, high-dimensional data, and evolving fraud patterns. A dataset comprising inpatient claims, outpatient claims, and beneficiary details was used to train and evaluate five ML models: Random Forest, KNN, LDA, Decision Tree, and AdaBoost. Data preprocessing techniques included resampling SMOTE method to address the class imbalance, feature selection for dimensionality reduction, and aggregation of diagnostic and procedural codes. Random Forest emerged as the best-performing model, achieving a training accuracy of 99.2% and validation accuracy of 98.8%, and F1-score (98.4%). The Decision Tree also performed well, achieving a validation accuracy of 96.3%. KNN and AdaBoost demonstrated moderate performance, with validation accuracies of 79.2% and 81.1%, respectively, while LDA struggled with a validation accuracy of 63.3% and a low recall of 16.6%. The results highlight the importance of advanced resampling techniques, feature engineering, and adaptive learning in detecting Medicare fraud effectively. This study underscores the potential of machine learning in addressing the complexities of fraud detection. Future work should explore explainable AI and hybrid models to improve interpretability and performance, ensuring scalable and reliable fraud detection systems that protect healthcare resources and beneficiaries.

## 1. Introduction

Medicare fraud poses a serious threat to the sustainability and integrity of healthcare systems, especially within large programs like Medicare in the United States, which provides healthcare coverage to millions. Fraudulent activities result in billions of dollars in financial losses each year, eroding trust in the healthcare system and undermining the quality of care provided to legitimate beneficiaries [1-3]. Detecting and preventing Medicare fraud is, therefore, a crucial focus for policymakers and healthcare providers alike. As fraudulent schemes become more sophisticated, traditional detection methods struggle to keep pace, but advances in technology particularly in machine learning are providing promising new solutions. Medicare fraud includes a range of deceptive practices such as billing for services not provided, overcharging, upcoding (claiming higher-cost services than those actually performed), and using false diagnoses to inflate claims [3]. Identifying fraud within the enormous volume of legitimate claims processed daily is challenging. According to Herland et al. [1], the successful detection of Medicare fraud could potentially recover up to $350 billion in losses, emphasizing the critical need for effective fraud detection systems. However, this is easier said than done, as fraudulent claims make up only a tiny fraction of the total transactions, making detection akin to finding a needle in a haystack. Several key challenges arise in the effort to detect Medicare fraud effectively, many of which the proposed machine learning method aims to address.

One of the most significant challenges in Medicare fraud detection is the extreme class imbalance between fraudulent and non-fraudulent claims [2]. Fraudulent transactions account for less than 1% of total claims, which means that traditional machine learning models often fail to detect these rare instances or produce a high number of false positives. Standard algorithms tend to be biased toward the majority class, which in this case are legitimate claims, and struggle to identify fraudulent cases accurately[2]. The method addresses this challenge through advanced resampling techniques such as oversampling of fraudulent cases (e.g., SMOTE) and under sampling of legitimate ones, as well as hybrid techniques that maintain model performance while dealing with imbalanced data[3] [4]. Medicare claims data is often high-dimensional and highly structured, with hundreds of features such as patient demographics, provider information, diagnoses, and procedures. Extracting relevant features while filtering out irrelevant noise is critical to improving model performance. The approach incorporates feature selection and dimensionality reduction techniques to streamline the data and highlight the most important indicators of fraud [5]. By refining the data fed into machine learning models, increase the likelihood of accurately detecting fraudulent claims without overwhelming the model with unnecessary complexity [6]. Fraudulent behaviors evolve as fraudsters develop new ways to exploit the Medicare system. Static models often struggle to keep pace with these changes. To combat this, the method incorporates machine learning models that can adapt over time. By continuously retraining on updated data, the approach ensures that the detection system evolves alongside emerging fraud patterns, reducing the risk of outdated models missing new types of fraud[5]. This dynamic approach to fraud detection enables us to identify evolving fraudulent activities more effectively than traditional static models. While identifying fraudulent claims is essential, it is equally important to minimize the occurrence of false positives cases where legitimate claims are incorrectly flagged as fraud. High false-positive rates can result in unnecessary investigations and strain on healthcare providers[6]. The proposed solution uses precision-tuned algorithms that optimize for both sensitivity (identifying fraud) and specificity (reducing false positives), ensuring that the detection system is both accurate and practical for real-world applications. Given the size and complexity of Medicare datasets, scalability is a major concern. Fraud detection models must be able to handle millions of claims in real-time without sacrificing accuracy or speed. The method leverages deep learning models, which are particularly adept at processing large datasets, and introduces techniques such as parallel computing and distributed processing to ensure that the model can scale efficiently while maintaining high performance[7].

In this paper, we present a comprehensive framework for Medicare fraud detection using advanced ML techniques to address critical challenges, including class imbalance, high-dimensional data, and the dynamic nature of fraudulent schemes. This study makes several key contributions to the field of fraud detection in healthcare systems. First, we tackle the issue of extreme class imbalance, where fraudulent claims represent less than 1% of the total data. Using resampling techniques such as Synthetic Minority Oversampling Technique (SMOTE), we balance the dataset effectively, ensuring that ML models are sensitive to minority class patterns while minimizing overfitting. This approach improves the detection of rare fraudulent claims, which is often overlooked by standard models. Second, we address the high-dimensional nature of Medicare datasets by implementing feature selection and dimensionality reduction techniques. By streamlining the dataset, we preserve key indicators of fraud, such as suspicious billing patterns and procedural anomalies, while reducing computational complexity. This step enhances model efficiency and interpretability, enabling better performance on both training and validation datasets. Third, we integrate adaptive learning mechanisms into our ML models to ensure they remain effective against evolving fraud patterns. By continuously retraining models on updated datasets, we provide a dynamic solution that adapts to new fraudulent behaviors, outperforming static detection methods. Fourth, we conduct a comparative evaluation of multiple ML algorithms, including Random Forest, Decision Tree, KNN, LDA, and AdaBoost, to identify the most effective model. Our results demonstrate that Random Forest and Decision Tree achieve superior performance in detecting fraudulent claims, with high accuracy, precision, and recall. From a business perspective, the framework introduces transformative advancements by significantly reducing financial losses through enhanced fraud detection accuracy while maintaining operational efficiency with scalable and adaptive solutions. By addressing class imbalance, leveraging advanced feature selection, and incorporating dynamic learning, it enables real-time fraud detection, reduces false positives, and minimizes unnecessary investigations, thereby saving costs and improving trust in healthcare operations. The system's explainability fosters transparency for decision-makers, while its scalability and predictive insights offer a proactive approach to fraud management. Additionally, this innovative methodology provides a competitive edge, ensures regulatory compliance, and holds potential for broader applications in other fraud-prone industries, making it a valuable asset for organizations aiming to enhance resilience and efficiency.

Overall, this paper advances Medicare fraud detection by proposing a robust, scalable, and adaptive ML-based framework that improves detection accuracy, minimizes false positives, and ensures practical applicability in real-world healthcare environments. From a theoretical perspective, this paper provides significant contributions to the field of Medicare fraud detection by addressing critical challenges through advanced machine learning techniques and proposing a structured framework for systematic detection. Theoretically, it tackles the issue of extreme class imbalance, introducing and validating resampling methods such as SMOTE to improve model sensitivity to minority class patterns. It also contributes to the understanding of high-dimensional data processing by applying feature selection and dimensionality reduction techniques, which preserve key fraud indicators while minimizing computational complexity. Furthermore, the integration of adaptive learning mechanisms highlights a novel approach to handling the evolving nature of fraudulent schemes, ensuring model robustness over time. The comparative evaluation of multiple machine learning algorithms, including Random Forest, Decision Tree, KNN, LDA, and AdaBoost, provides a theoretical basis for selecting optimal models based on accuracy, precision, recall, and scalability. Overall, the paper advances theoretical knowledge in the application of machine learning to healthcare fraud detection, emphasizing the importance of balancing detection accuracy, computational efficiency, and adaptability.

This paper is organized into five main sections. The introduction provides an overview of Medicare fraud, its challenges, and the potential of machine learning in addressing these issues. The literature review discusses prior research on fraud detection, emphasizing class imbalance, feature selection, and adaptive learning. The methods and materials section outlines the data preprocessing, feature engineering, resampling techniques, and machine learning models used. The results and discussion section presents the evaluation of model performance, highlighting key findings and comparisons. Finally, the conclusion summarizes the contributions, numerical results, and implications, while suggesting directions for future research in fraud detection.

## 2. Literature review

Medicare fraud detection is a critical issue that requires ongoing attention from researchers, policymakers, and healthcare providers [3-5]. The use of advanced machine learning models, combined with techniques for addressing class imbalance and dimensionality reduction, has significantly improved the ability to detect fraudulent activities in Medicare claims. However, challenges remain, particularly in the areas of model evaluation, real-world validation, and the ethical use of machine learning for fraud detection. As healthcare systems continue to evolve, so too must the methods used to protect them from fraud, ensuring that resources are directed toward providing high-quality care for legitimate beneficiaries.

### 2.1. The Challenges of Class Imbalance in Medicare Fraud Detection

One of the most pressing challenges in Medicare fraud detection is the issue of class imbalance, where the number of legitimate claims vastly outweighs the number of fraudulent ones. As Bauder et al. [8] highlighted, in a dataset where only 0.062% of providers were identified as fraudulent, the severe class imbalance poses a significant problem for machine learning models. Most models, when trained on such datasets, tend to be biased toward the majority class, resulting in poor performance when it comes to identifying fraud. To address this, various techniques have been employed. Bounab et al.[9] proposed a hybrid method combining the SMOTE with Edited Nearest Neighbor (ENN) to mitigate class imbalance in Medicare fraud detection. Their approach improved detection accuracy by addressing the limitations of traditional oversampling methods, such as overfitting and noise generation. These findings underscore the importance of using advanced sampling techniques to balance datasets and ensure that models can accurately detect fraudulent activities.

### 2.2. The Role of Machine Learning in Medicare Fraud Detection

Machine learning has emerged as one of the most promising tools for detecting fraud in Medicare claims. The ability of machine learning algorithms to process large volumes of data and identify patterns makes them well-suited for this task. Over the years, numerous machine learning models have been developed and applied to Medicare fraud detection, each with its strengths and weaknesses. For example, Johnson et al.[10] explored the impact of class label noise on detecting fraud within highly imbalanced datasets. They evaluated four popular machine learning algorithms, including deep learning models, and found that noisy labels significantly affect the accuracy of fraud detection models. Their study highlights the need for robust data preprocessing and label cleaning to improve the performance of fraud detection systems. Another key development in machine learning for fraud detection is the use of neural networks. Johnson et al. [11] applied deep neural networks to Medicare fraud detection and found that random oversampling (ROS) techniques improved model performance. Similarly, Mayaki et al. [12] developed a deep neural network with an autoencoder component to predict fraudulent claims in Medicare. Their model, which considered multiple data sources, showed significant improvements in fraud detection accuracy. The versatility of neural networks, combined with their ability to process complex and high-dimensional data, makes them particularly useful in this domain.

### 2.3. Graph-Based Approaches to Fraud Detection

In addition to traditional machine learning models, graph-based approaches have shown considerable promise in Medicare fraud detection. Yoo et al. [13] explored the use of graph neural networks (GNNs) to detect fraud in Medicare claims by analyzing the relationships between healthcare providers, beneficiaries, and services. By converting Medicare data into a graph structure, they were able to leverage the interconnectedness of these entities to improve fraud detection. Their study found that GNN-based models outperformed traditional machine learning models, demonstrating the potential of graph-based methods in identifying fraudulent activities. Graph-based approaches are particularly effective because they can model the complex relationships that exist in Medicare data. Fraudulent providers often interact with multiple entities, and their activities can be identified by analyzing these relationships. For instance, a provider who frequently bills for services that are not typically required by a specific patient population may be flagged as suspicious. By leveraging the power of graph analytics, Medicare fraud detection systems can uncover hidden patterns that may not be apparent in traditional tabular datasets.

### 2.4. The Impact of Feature Selection and Dimensionality Reduction

One of the challenges in Medicare fraud detection is the high dimensionality of the data. Medicare claims datasets often contain thousands of features, many of which may be irrelevant or redundant. To improve the efficiency and accuracy of fraud detection models, feature selection and dimensionality reduction techniques are often employed. Wang et al.[14] addressed this issue by applying supervised feature selection methods within ensemble techniques. Their approach reduced the dimensionality of the dataset while preserving important features, leading to improved classification accuracy. Similarly, Johnson et al.[15] explored encoding high-dimensional procedure codes in healthcare fraud detection. They compared traditional one-hot encoding techniques with aggregation methods and found that using advanced encoding techniques, such as binary tree decomposition and hashing, significantly improved model performance. These studies demonstrate the importance of feature selection and dimensionality reduction in handling large and complex Medicare datasets.

### 2.5. Addressing Model Evaluation and Validation in Real-World Applications

An important consideration in Medicare fraud detection is the evaluation and validation of machine learning models in real-world settings. As Bauder et al.[16] pointed out, traditional cross-validation methods may not be sufficient for evaluating the performance of fraud detection models. Instead, it is crucial to test these models on new, unseen data to ensure their generalizability. Bauder et al. emphasized the importance of validating models on real-world Medicare claims data, as this provides a more accurate assessment of their effectiveness in detecting fraud. Furthermore, Levy et al.[17] explored the use of one-class classification versus binary classification for fraud detection in highly imbalanced datasets. Their study found that one-class classification outperformed binary classification, particularly when dealing with datasets where the majority of transactions are legitimate. These findings suggest that alternative evaluation methods, such as one-class classification, may be more appropriate for Medicare fraud detection due to the inherent imbalance in the data.

### 2.6. The Role of Predictive Analytics in Healthcare

Beyond fraud detection, predictive analytics plays a broader role in the healthcare sector, influencing decision-making processes for both providers and patients. Sharma et al.[18] investigated the impact of predictive analytics on healthcare, finding that these tools can help anticipate future health issues, plan treatments, and mitigate risks. The ability to forecast patient outcomes not only improves the quality of care but also aids in detecting fraudulent activities, as providers who engage in unusual billing practices may be identified more easily. Predictive analytics can also be applied to detect patterns in fraudulent behavior over time. As Levy et al.[19] highlighted, the performance of fraud detection models can degrade over time due to changes in data distributions. By continuously updating models and incorporating new data, healthcare organizations can ensure that their fraud detection systems remain effective. This dynamic approach to fraud detection is essential in a constantly evolving healthcare landscape, where new forms of fraud may emerge as providers and patients adapt to regulatory changes.

### 2.7. The Economic and Social Impact of Medicare Fraud

The economic impact of Medicare fraud is immense. According to estimates, Medicare fraud costs the U.S. government billions of dollars each year. This not only results in financial losses but also places a strain on the healthcare system, leading to higher premiums for patients and reduced resources for legitimate healthcare providers [1]. Mayaki et al.[12] noted that Medicare fraud results in higher premiums for clients, which can further exacerbate disparities in healthcare access and affordability. Medicare fraud also has significant social implications. When fraudulent providers drain resources from the system, it affects the quality of care that legitimate beneficiaries receive. For example, if a provider is billing for services that were not rendered, patients may not receive the necessary treatments, leading to poorer health outcomes. Furthermore, fraudulent activities can undermine trust in the healthcare system, making it more difficult for providers to deliver high-quality care.

Table 1: Summary of key studies on Medicare fraud detection

| Author | Year | Aim | Result |
|---|---|---|---|
| Bounab et al.[9] | 2024 | Proposed a hybrid method combining SMOTE with ENN to address data imbalance in healthcare fraud detection. | SMOTE-ENN with XGBoost improved efficiency in detecting fraud, outperforming traditional ML techniques. |
| Chirchi et al.[20] | 2024 | Studied healthcare fraud detection using advanced ML models to predict provider fraud in Medicare. | SMOTE and advanced models improved fraud detection accuracy and addressed class imbalance issues. |
| Yoo et al.[13] | 2023 | Investigated Medicare fraud detection using graph analysis to improve detection accuracy. | Graph neural networks outperformed traditional ML models in detecting Medicare fraud. |
| Wang et al.[14] | 2023 | Tackled high dimensionality and class imbalance in Medicare fraud detection using feature selection and RUS. | Feature selection improved classification accuracy, addressing class imbalance and high dimensionality. |
| Johnson and Khoshgoftaar[10] | 2023 | Introduced a data-centric approach to improve healthcare fraud classification performance using Medicare claims data. | Constructed large labeled datasets from CMS data, improving fraud detection reliability. |
| Hancock et al.[21] | 2023 | Developed explainable ML models for Medicare fraud detection using feature selection techniques. | Feature selection reduced dimensionality while maintaining accuracy, improving transparency in fraud detection. |
| Mayaki et al.[12] | 2022 | Examined the use of neural networks with autoencoders to predict fraudulent Medicare claims. | Deep neural network architecture improved classification accuracy for detecting Medicare fraud. |
| Matschak et al.[22] | 2022 | Presented a CNN-based approach for detecting health insurance claim fraud. | Achieved an AUC of 0.7 for selected fraud types using a CNN-based approach. |

| Kumaraswamy et al.[23] | 2022 | Reviewed data mining methods for healthcare fraud detection, focusing on digital systems. | Highlighted challenges in implementing digital fraud detection systems in healthcare. |
|---|---|---|---|
| Johnson et al.[15] | 2022 | Studied encoding high-dimensional procedure codes for healthcare fraud detection. | Binary tree decomposition and hashing improved classification accuracy for fraud detection. |
| Bauder & Khoshgoftaar[24] | 2020 | Investigated the effects of class rarity on binary classification problems in Big Data fraud detection. | Oversampling and undersampling techniques improved model accuracy and reduced bias. |
| Johnson et al.[25] | 2020 | Explored the influence of medical provider specialty and semantic embeddings in detecting fraudulent providers. | Dense semantic embeddings improved model performance for detecting fraudulent providers. |
| Arunkumar et al.[26] | 2021 | Investigated hybrid clustering and classification methods for healthcare insurance fraud detection. | Hybrid clustering and classification approaches outperformed other algorithms in classifying fraud. |
| Johnson et al.[11] | 2019 | Applied neural networks to detect Medicare fraud with publicly available claims data. | Improved existing ML models for Medicare fraud detection, contributing to automated fraud detection. |
| Bauder et al.[16] | 2019 | Evaluated ML model performance for real-world Medicare fraud detection. | Stressed the importance of validating ML models on new input data for real-world applications. |
| Johnson et al.[11] | 2019 | Explored the use of thresholding with deep neural networks for class imbalance in Medicare fraud detection. | ROS outperformed undersampling and SMOTE for handling class imbalance in Medicare fraud detection. |
| Obodoekwe et al.[27] | 2019 | Compared ML methods for detecting healthcare claims fraud. | Ensemble methods and neural networks were the most effective, while logistic regression performed poorly. |
| Hasanin et al.[28] | 2019 | Analyzed the impact of class imbalance on Big Data analytics using ML algorithms. | Emphasized the importance of data sampling strategies to mitigate class imbalance in Big Data. |
| Herland et al.[1] | 2019 | Investigated class rarity in supervised healthcare fraud detection models using Medicare data. | Detected fraudulent activities could recover up to $350 billion in financial losses. |

| Herland et al.[29] | 2017 | Developed an anomaly detection model to identify potential healthcare fraud. | Improved anomaly detection model performance through feature selection and specialty grouping. |
| --- | --- | --- | --- |

Table 1 show the summary of key studies on Medicare fraud detection, highlighting the aims and results of various approaches. The studies range from investigating class imbalance and model performance to exploring new machine learning techniques such as SMOTE-ENN and GraphSAGE for improving fraud detection accuracy in highly imbalanced datasets.

## 3. Methods and Material

The process for detecting Medicare fraud involves a structured sequence of steps, as illustrated in the Figure1. These steps include data preprocessing, feature engineering, dataset splitting, and model implementation. The overall approach incorporates the use of multiple machine learning algorithms to identify fraudulent claims effectively. Data preprocessing begins by preparing the raw dataset for analysis. Categorical variables, such as chronic condition indicators, are converted into numerical formats to facilitate machine learning. For example, values like "Y" are replaced with binary representations, while multi-class categorical variables are transformed using one-hot encoding. Missing data in numerical fields, such as admission dates and deductible amounts, is handled using imputation techniques. Domain-specific logic is applied to replace missing values with zeros or calculated averages. To reduce noise in the dataset, features with minimal relevance to fraud detection are excluded. Financial data, including claim amounts and reimbursements, is normalized to ensure consistency and comparability across records. Additionally, diagnosis and procedure codes are aggregated into indexed values or groups, preserving patterns while reducing the dataset's dimensionality. Feature engineering further enhances the dataset by creating refined variables that capture underlying relationships. For numeric features, averages of key metrics such as insurance claims and reimbursement amounts are calculated to represent overall trends in provider behavior. Diagnosis and procedure codes are grouped into indices, simplifying the complexity of these fields while retaining their informative value. Once feature engineering is complete, the dataset is split into training, validation, and testing subsets. Given the imbalance in the dataset, with fraudulent claims being relatively rare, balancing techniques such as the SMOTE are applied to ensure that the machine learning models are not biased toward the majority class. Model implementation involves the application of multiple machine learning algorithms to identify fraudulent patterns in claims. The target variable is converted into binary labels, indicating whether a claim is fraudulent or not. Several machine learning models, including Random Forest, KNN, LDA, DT, and AdaBoost, are trained to capture complex relationships between features. The models are validated on the test dataset to evaluate their effectiveness. Performance metrics such as accuracy, precision, recall, and F1-score are calculated to assess the models and determine the most suitable algorithm for the task.

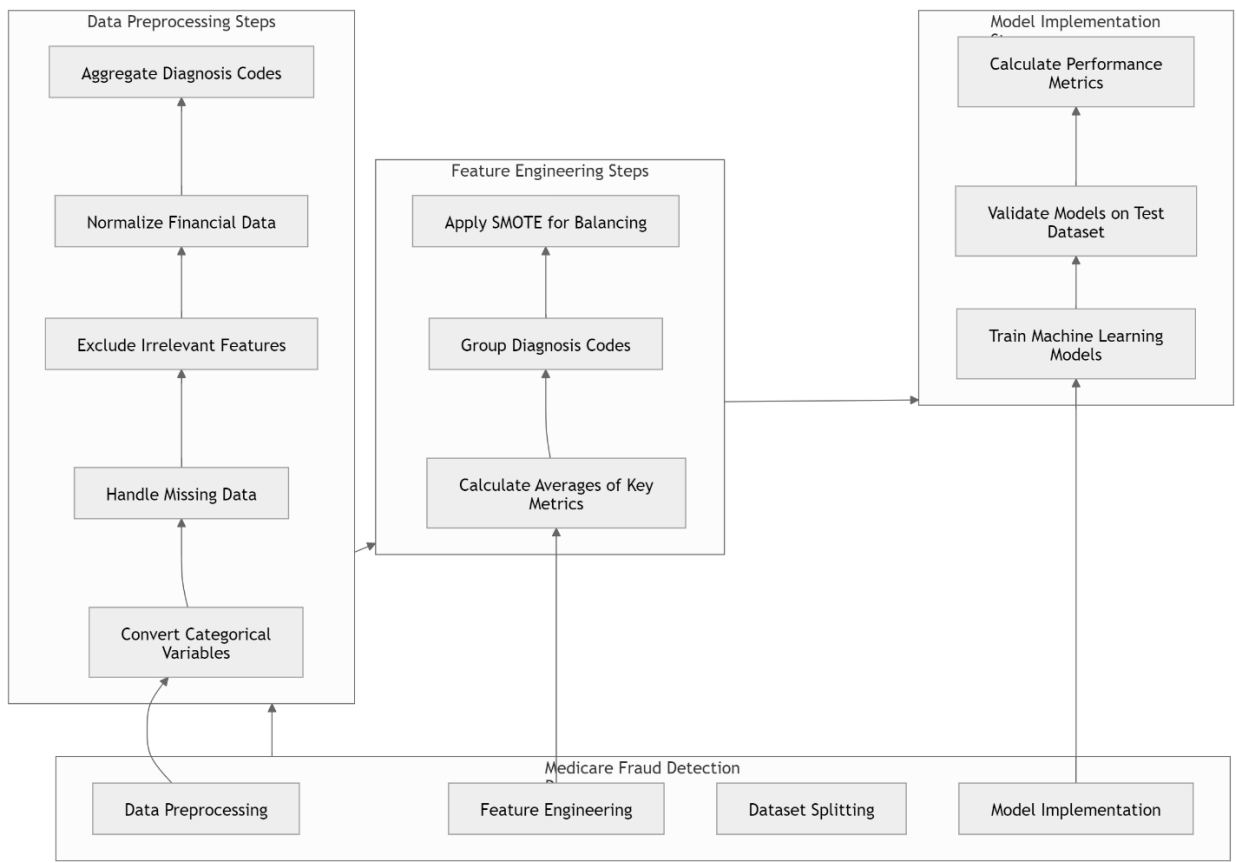

**Figure 1:** Workflow of the Medicare Fraud Detection Process, detailing data preprocessing, feature engineering, dataset balancing, and the implementation of multiple machine learning models for fraud identification and evaluation.

Figure 1 provides a visual representation of this methodology. It starts with data preprocessing, where categorical variables are transformed, irrelevant features are removed, and missing data is imputed. This ensures that the dataset is clean and consistent. Feature engineering follows, where diagnosis and procedure codes are aggregated, and averages are calculated for numerical features. The dataset is then split into subsets for training, validation, and testing. Balancing techniques like SMOTE are used to address class imbalance, ensuring that fraudulent claims are adequately represented in the training data. The final stage involves training multiple machine learning algorithms and evaluating their performance. Each model's output is analyzed using key evaluation metrics to select the best-performing approach. The iterative process and structured methodology for Medicare fraud detection are detailed in the Methods and Materials section, where the evaluation of multiple machine learning models, including Random Forest, Decision Tree, KNN, LDA, and AdaBoost, is performed using metrics like accuracy, precision, recall, and F1-score to identify the most effective approach. This process is visually represented in the Figure 1, which outlines the sequential steps: data preprocessing, feature engineering, dataset balancing, and model implementation. Data preprocessing involves transforming categorical variables, handling missing data, and normalizing financial variables, while feature engineering focuses on aggregating diagnostic and procedural codes to refine the dataset. Resampling techniques like SMOTE address class imbalance, ensuring that the models are sensitive to minority class patterns. This systematic and iterative approach ensures that the chosen model is robust, scalable, and well-suited for real-world application in identifying anomalies and fraudulent activities in Medicare claims. As Medicare fraud continues to evolve, so too must the methods used to detect and prevent it. Future research in this field should focus on the development of more sophisticated machine learning models that can handle the complexities of Medicare data. In particular, researchers should explore the use of hybrid models that combine the strengths of different machine learning techniques. For example, combining deep learning with graph-based approaches could yield powerful fraud detection systems that can uncover hidden patterns in Medicare claims data. Additionally, more attention should be given to the ethical implications of using machine learning for fraud detection. While these models can significantly improve fraud detection accuracy, there is a risk of false positives, where legitimate providers are wrongly accused of fraud. Policymakers and researchers must work together to ensure that fraud detection systems are fair and transparent, with mechanisms in place to address any potential biases [30].

## 4. Results and Discussion

### 4.1. Data collection

The dataset used in this study was curated to analyze fraudulent behavior in Medicare claims. It consists of three distinct data sources: Inpatient claims, Outpatient claims, and Beneficiary details. These sources collectively provide a view of provider claims, patient admissions, and associated financial transactions. Each dataset was obtained from anonymized records to maintain confidentiality and ensure compliance with ethical standards in data handling. The inpatient claims data represents hospital admission-related transactions, encompassing patients who were formally admitted to healthcare facilities[31]. This dataset includes fields such as Admission Date, Discharge Date, and Admission Diagnosis Code. These variables enable the identification of patterns in the length of stay, diagnosis consistency, and claim frequency for individual providers. For example, an unusually high frequency of admissions for specific diagnoses could indicate potential upcoding or billing for unnecessary services. Moreover, fields like ClmDiagnosisCode_1 to ClmDiagnosisCode_10 provide detailed information about diagnoses associated with each claim, which can be used to cross-verify the alignment between the claimed diagnosis and the performed procedures. The outpatient claims data captures information about services provided to patients who were not admitted to the hospital. This dataset includes fields such as Provider ID, Claim Start Date, Claim End Date, and Claim Diagnosis Codes. These variables are used for detecting overutilization of services, duplicate claims, or instances where the level of service billed exceeds what was performed. For example, the presence of duplicate Claim IDs or a mismatch between Claim Diagnosis Codes and Procedure Codes may suggest intentional misrepresentation by a provider. The beneficiary dataset contains information related to patient demographics and health conditions, including variables such as BeneID, Deductible Amount Paid, and Region Code. These fields enable profiling of patients and their healthcare needs, which can be contrasted against the claims submitted by providers. For instance, if certain regions show a disproportionately high frequency of claims for specific providers, it could signal collusion or localized fraudulent activities. Additionally, tracking variables such as Deductible Amount Paid allows for identifying unusual financial patterns that may correlate with fraudulent behavior. Table 2 provides an overview of variables included in the dataset, categorized by beneficiary details, claims information, physician involvement, diagnosis codes, procedure codes, and financial details. Each variable is described to highlight its relevance in identifying potential fraudulent activities in Medicare claims.

**Table 2**: Detailed List of Variables in the Medicare Fraud Dataset

| Variable Name | Description |
|---|---|
| BeneID | Unique identifier for each beneficiary. |
| DOB | Date of birth of the beneficiary. |
| DOD | Date of death of the beneficiary. |
| Gender | Gender of the beneficiary. |
| Race | Race of the beneficiary. |
| RenalDiseaseIndicator | Indicator if the beneficiary has renal disease. |
| State | State code where the beneficiary resides. |
| County | County code where the beneficiary resides. |
| NoOfMonths_PartACov | Number of months the beneficiary was covered under Medicare Part A. |
| NoOfMonths_PartBCov | Number of months the beneficiary was covered under Medicare Part B. |
| ChronicCond_Alzheimer | Indicator if the beneficiary has Alzheimer's disease. |
| ChronicCond_Heartfailure | Indicator if the beneficiary has heart failure. |
| ChronicCond_KidneyDisease | Indicator if the beneficiary has kidney disease. |
| ChronicCond_Cancer | Indicator if the beneficiary has cancer. |
| ChronicCond_ObstrPulmonary | Indicator if the beneficiary has obstructive pulmonary disease. |
| ChronicCond_Depression | Indicator if the beneficiary has depression. |
| ChronicCond_Diabetes | Indicator if the beneficiary has diabetes. |
| ChronicCond_IschemicHeart | Indicator if the beneficiary has ischemic heart disease. |
| ChronicCond_Osteoporasis | Indicator if the beneficiary has osteoporosis. |
| ChronicCond_RheumatoidArthritis | Indicator if the beneficiary has rheumatoid arthritis. |

| ChronicCond_Stroke | Indicator if the beneficiary has experienced a stroke. |
|---|---|
| IPAnnualReimbursementAmt | Annual inpatient reimbursement amount. |
| IPAnnualDeductibleAmt | Annual inpatient deductible amount. |
| OPAnnualReimbursementAmt | Annual outpatient reimbursement amount. |
| OPAnnualDeductibleAmt | Annual outpatient deductible amount. |
| ClaimID | Unique identifier for each claim. |
| ClaimStartDt | Start date of the claim. |
| ClaimEndDt | End date of the claim. |
| Provider | Unique identifier for the healthcare provider. |
| InscClaimAmtReimbursed | Insurance claim amount reimbursed. |
| AttendingPhysician | Identifier for the attending physician. |
| OperatingPhysician | Identifier for the operating physician. |
| OtherPhysician | Identifier for any other physician linked to the claim. |
| ClmDiagnosisCode_1 to 10 | Up to 10 diagnosis codes for a single claim. |
| ClmProcedureCode_1 to 6 | Up to 6 procedure codes for medical procedures performed during the claim period. |
| DeductibleAmtPaid | Deductible amount paid by the beneficiary for the claim. |
| ClmAdmitDiagnosisCode | Admission diagnosis code for inpatient claims. |

### *4.2. Findings*

The violin plots in Figure 2 depict the statistical distribution of key inpatient-related variables from the Medicare dataset, offering insights into claim characteristics and potential anomalies. In Figure 2 (a), the "Number of Days Admitted" variable shows a heavily right-skewed distribution. The median length of stay is approximately 4–6 days, with the interquartile range (IQR) primarily falling below 10 days. Rare outliers extend beyond 30 days, indicating unusually prolonged hospitalizations. These outliers could correspond to complex medical cases, but they also raise concerns about potentially exaggerated claims for prolonged services. The density plot shows a sharp peak around shorter stays, aligning with typical inpatient scenarios for standard medical conditions. Figure 2 (b) highlights the "Insurance Claim Amount Reimbursed." The majority of claims fall below $20,000, with a sharp decline in frequency as reimbursement amounts increase. The median reimbursed amount is approximately $5,000–$8,000. A long tail in the distribution, extending beyond $100,000, suggests a small subset of high-cost claims.

These high-value reimbursements may represent cases of upcoding (assigning more expensive billing codes than appropriate) or billing for unnecessary services, both of which are common fraudulent patterns in healthcare.

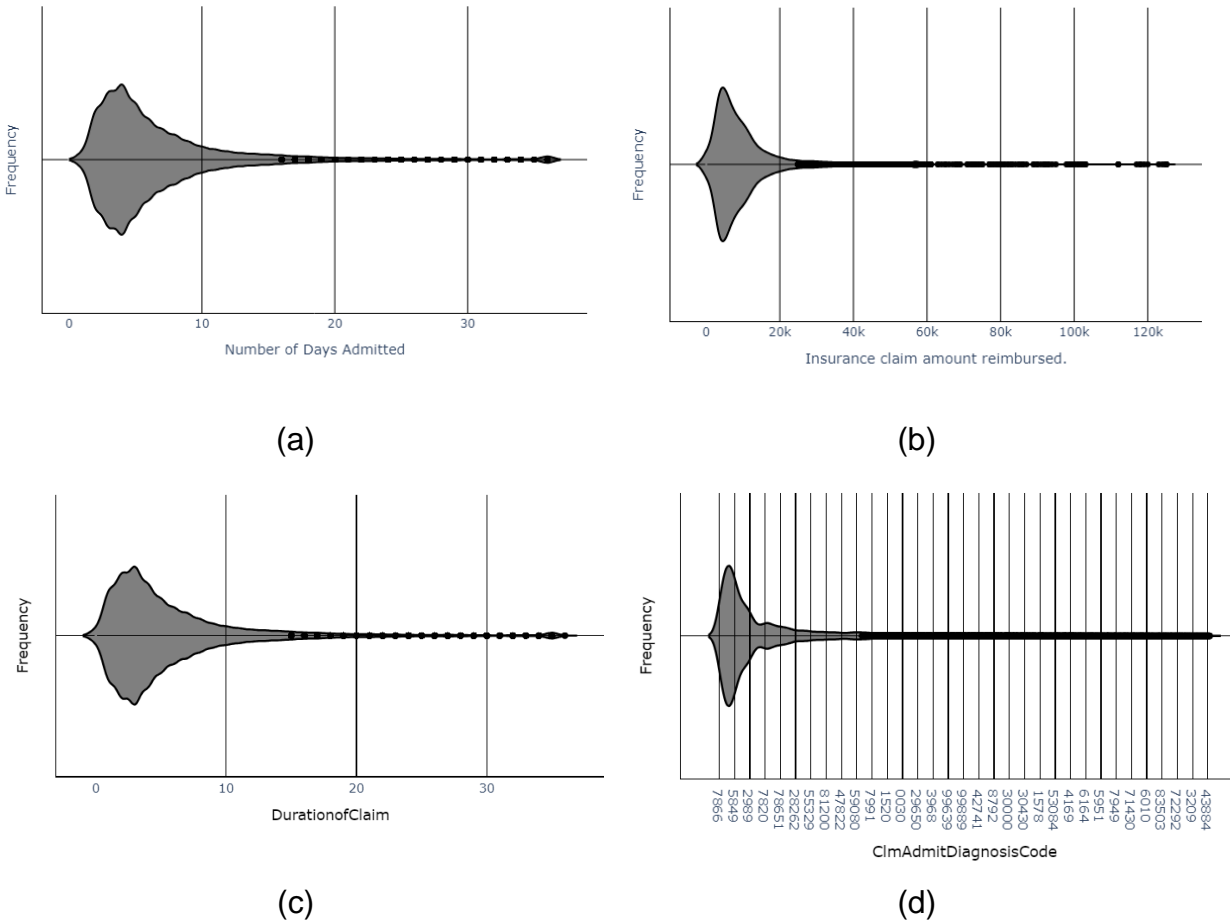

(a)

(b)

(c)

(d)

**Figure 2:** Violin plots representing the distribution of inpatient claim variables in the Medicare dataset: (a) Number of Days Admitted, showing the prevalence of short stays with rare prolonged hospitalizations; (b) Insurance Claim Amount Reimbursed, highlighting a skewed distribution with a small proportion of high-value claims; (c) Duration of Claim, indicating most claims are resolved within 10 days with occasional outliers; and (d) Admission Diagnosis Codes, demonstrating the frequent use of specific codes and the presence of less common diagnoses.

In Figure 2 (c), the "Duration of Claim" variable exhibits a similar distribution to the length of stay. Claims typically close within 5–10 days, as indicated by the dense clustering in this range. Outliers beyond 30 days point to potential anomalies such as delayed processing, unresolved cases, or deliberate manipulation to increase payouts. The consistency in short durations aligns with standard hospital billing practices for inpatient services. Figure 2 (d) represents the distribution of "Admission Diagnosis Codes," which are categorical variables converted into numerical indices for analysis. A subset of diagnosis codes appears to dominate the dataset, with a small number of codes accounting for the majority of admissions.

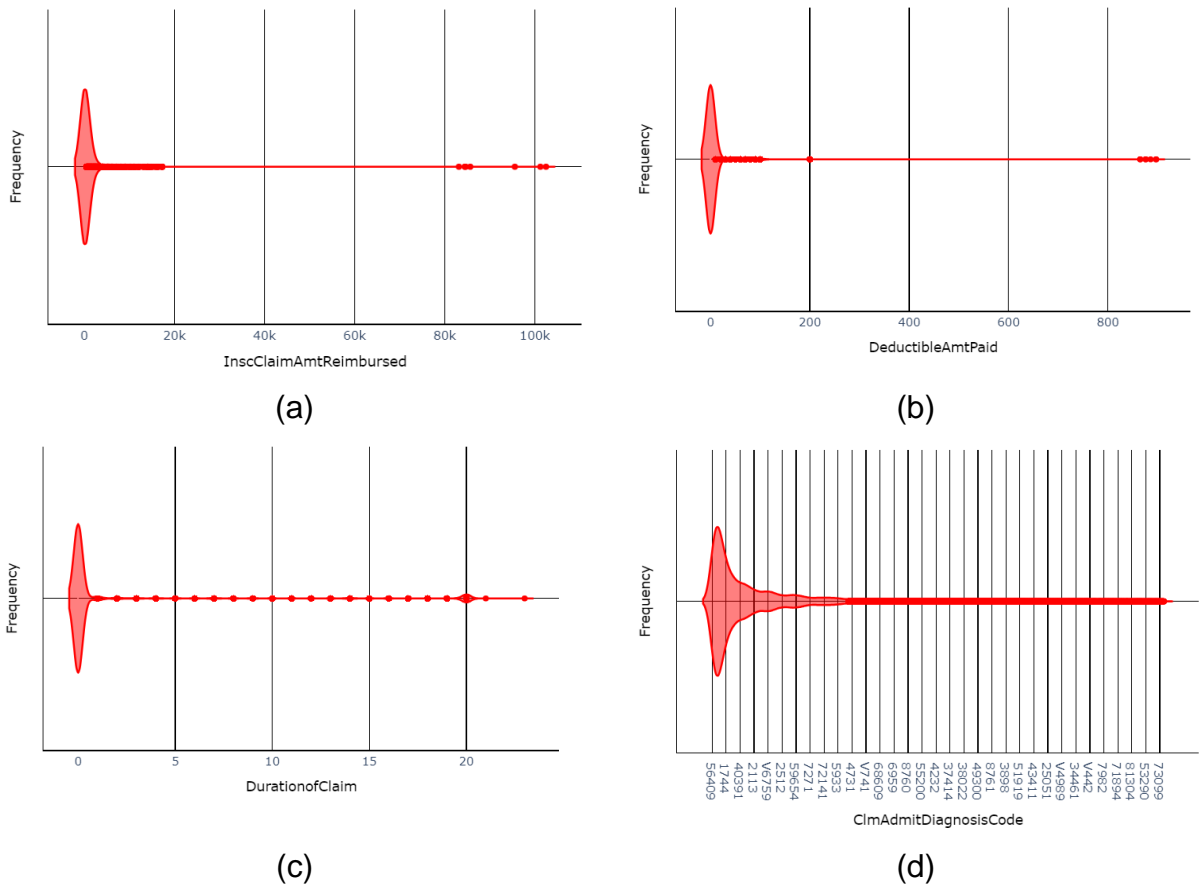

(a)

(b)

(c)

(d)

**Figure 3:** Violin plots for outpatient claim variables in the Medicare dataset: (a) Insurance Claim Amount Reimbursed, showing a skewed distribution with rare high-value claims; (b) Deductible Amount Paid, concentrated below $200 with occasional high outliers; (c) Duration of Claim, reflecting short claim periods with few extended durations; and (d) Admission Diagnosis Codes, highlighting frequent use of specific codes with a long tail for less common diagnoses.

For example, codes frequently used at rates higher than expected may suggest provider-specific biases, such as overuse of certain diagnoses to justify admissions. Conversely, the long tail of less common codes likely corresponds to rarer medical conditions. Overrepresentation of specific codes can be further analyzed for patterns indicative of fraudulent coding practices.

The violin plots in Figure 3 illustrate the distribution of key outpatient claim variables in the Medicare dataset, highlighting patterns in financial, temporal, and diagnostic data. In Figure 3(a), the "Insurance Claim Amount Reimbursed" exhibits a skewed distribution, with the majority of reimbursements clustered below $20,000. The median reimbursement amount falls between $2,000 and $5,000, reflecting typical outpatient service costs. A long tail extending beyond $100,000 represents rare, high-value claims. These extreme values could indicate anomalies or potentially fraudulent activities, such as billing for unperformed services or improper upcoding. Figure 3(b) examines the "Deductible Amount Paid," which also shows a right-skewed pattern. Most deductible payments are below $200, with a median near $100. However, higher deductible payments exceeding $800 are infrequent and may correspond to cases involving complex outpatient procedures or financial irregularities. In Figure 3(c), the "Duration of Claim" shows most claims being resolved within 0 to 5 days. The distribution narrows significantly beyond 20 days, with very few claims extending further. Such prolonged claim durations could signify procedural delays or intentional manipulation to inflate billing periods. Figure 3(d) presents the distribution of "Admission Diagnosis Codes," where specific codes dominate outpatient claims. The heavy concentration of certain codes suggests common diagnoses in outpatient settings. However, repeated use of specific codes across claims may indicate systematic misrepresentation or overuse by providers.

The analysis reveals notable differences between inpatient and outpatient claim variables. Inpatient claims, represented by longer hospital stays and higher reimbursement amounts, show more variability, with frequent outliers in duration and cost. In contrast, outpatient claims are generally shorter in duration, with most reimbursements and deductible payments concentrated in lower ranges. While inpatient diagnosis codes reflect a broader variety of conditions, outpatient claims display a stronger dominance of specific codes. Outpatient claims exhibit fewer high-value outliers compared to inpatient claims, suggesting differing patterns of service intensity and cost, with the latter potentially carrying a higher risk of fraudulent activities.

**Table 3:** Descriptive statistics of key variables in the Medicare fraud, including counts, means, and standard deviations for potential fraud, reimbursement amounts, deductibles, admission rates, claim durations, and Medicare Part A and B coverage periods.

| Variable | Count | Mean | Std |
|---|---|---|---|
| Potential Fraud | 556,703 | 0.381 | 0.486 |
| Insurance Claim Amt Reimbursed | 556,703 | 996.936 | 3819.692 |
| Deductible Amt Paid | 556,703 | 78.428 | 273.809 |
| Admitted | 556,703 | 0.073 | 0.259 |
| Duration of Claim | 556,703 | 1.728 | 4.905 |
| Number of Days Admitted | 556,703 | 0.483 | 2.299 |
| Renal Disease Indicator | 556,703 | 0.197 | 0.397 |
| No. of Months Part A Cov | 556,703 | 11.931 | 0.890 |
| No. of Months Part B Cov | 556,703 | 11.939 | 0.786 |

The descriptive statistics summarize key variables in the Medicare fraud dataset, consisting of 556,703 records. The mean "Potential Fraud" rate is 0.381, indicating approximately 38% of claims are flagged as potentially fraudulent. The average "Insurance Claim Amount Reimbursed" is $996.94, with a high standard deviation of $3819.69, reflecting significant variability in claim amounts. The mean "Deductible Amount Paid" is $78.43, also showing substantial variation. Only 7.3% of claims involve admitted patients, with an average "Duration of Claim" of 1.73 days. Beneficiaries are covered under Part A and Part B for approximately 12 months on average, with minimal variability.

**Table 4:** Performance metrics (accuracy, precision, recall, and F1-score) of five machine learning models—Random Forest, KNN, LDA, Decision Tree, and AdaBoost—evaluated on both training and validation datasets for Medicare fraud detection.

| Model | Metric | Accuracy | Precision | Recall | F1-Score |
|---|---|---|---|---|---|
| Random Forest | Train | 0.992 | 0.985 | 1.000 | 0.992 |
| | Validation | 0.988 | 0.970 | 0.999 | 0.984 |
| KNN | Train | 0.895 | 0.853 | 0.955 | 0.901 |
| | Validation | 0.792 | 0.683 | 0.852 | 0.758 |
| LDA | Train | 0.665 | 0.841 | 0.408 | 0.549 |
| | Validation | 0.633 | 0.569 | 0.166 | 0.257 |
| Decision Tree | Train | 0.968 | 0.944 | 0.996 | 0.969 |
| | Validation | 0.963 | 0.912 | 1.000 | 0.954 |
| AdaBoost | Train | 0.836 | 0.816 | 0.866 | 0.840 |
| | Validation | 0.811 | 0.721 | 0.829 | 0.771 |

The results from the evaluation of five machine learning models—Random Forest, KNN, LDA, Decision Tree, and AdaBoost—demonstrate in Table 4 the varied performance of each in detecting Medicare fraud. Random Forest achieved the highest overall metrics, with a training accuracy of 99.2% and validation accuracy of 98.8%, accompanied by a near-perfect recall (99.9%) and F1-score (98.4%). Its strong validation performance highlights its ability to generalize effectively to unseen data. The Decision Tree also performed well, achieving a validation accuracy of 96.3% with precision and recall values exceeding 91%. Although slightly less robust than Random Forest, its simplicity makes it a competitive choice for this task. In contrast, KNN and AdaBoost showed moderate results. KNN achieved validation accuracy of 79.2% with a relatively lower precision (68.3%) and F1-score (75.8%). AdaBoost demonstrated better balance, with validation accuracy at 81.1% and a validation F1-score of 77.1%, but it still underperformed compared to Random Forest and Decision Tree. LDA struggled significantly, with a validation accuracy of only 63.3%, due to its low recall (16.6%), which indicates a poor ability to detect fraudulent cases. Overall, Random Forest and Decision Tree proved to be the most effective models for fraud detection in this study.

The bar charts in Figure 4 illustrate the training and validation metrics (accuracy, precision, recall, and F1-score) for five machine learning models applied to Medicare fraud detection. In Figure (a), which represents the training metrics, Random Forest and Decision Tree exhibit near-perfect scores across all metrics, indicating that these models effectively learned from the training data. KNN also performed well during training but showed slightly lower recall compared to the top-performing models. AdaBoost displayed balanced training metrics, though slightly lower than Decision Tree and Random Forest. LDA demonstrated the weakest performance in training, with a recall of only 40.8% and an F1-score of 54.9%, suggesting limited learning capacity on the provided data. Figure (b), representing validation metrics, highlights the generalization ability of the models. Random Forest maintains its strong performance, with accuracy and recall exceeding 98%, indicating robustness in detecting fraudulent claims. Decision Tree also performs well, achieving a validation accuracy of 96.3% and recall of 100%, though slightly lower precision affects its F1-score. KNN and AdaBoost show moderate performance, with validation accuracies of 79.2% and 81.1%, respectively. However, LDA struggles significantly on validation, with a recall of only 16.6%, indicating its inability to identify fraudulent cases effectively.

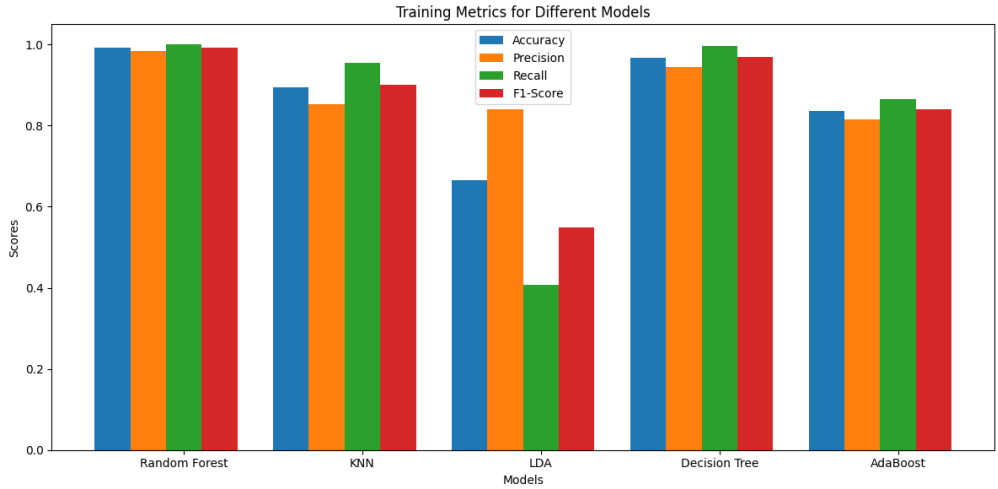

(a)

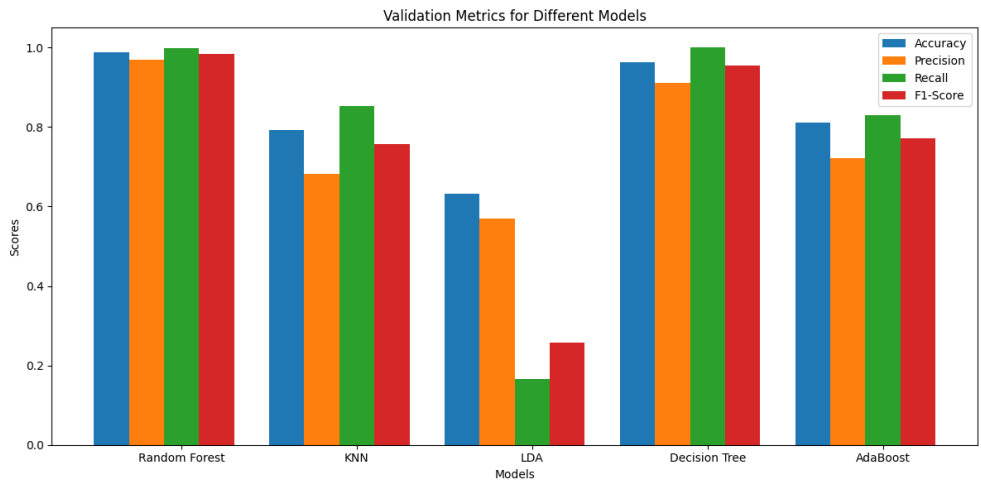

(b)

**Figure 4:** (a) Training metrics and (b) validation metrics (accuracy, precision, recall, F1-score) for Random Forest, KNN, LDA, Decision Tree, and AdaBoost models in Medicare fraud detection.

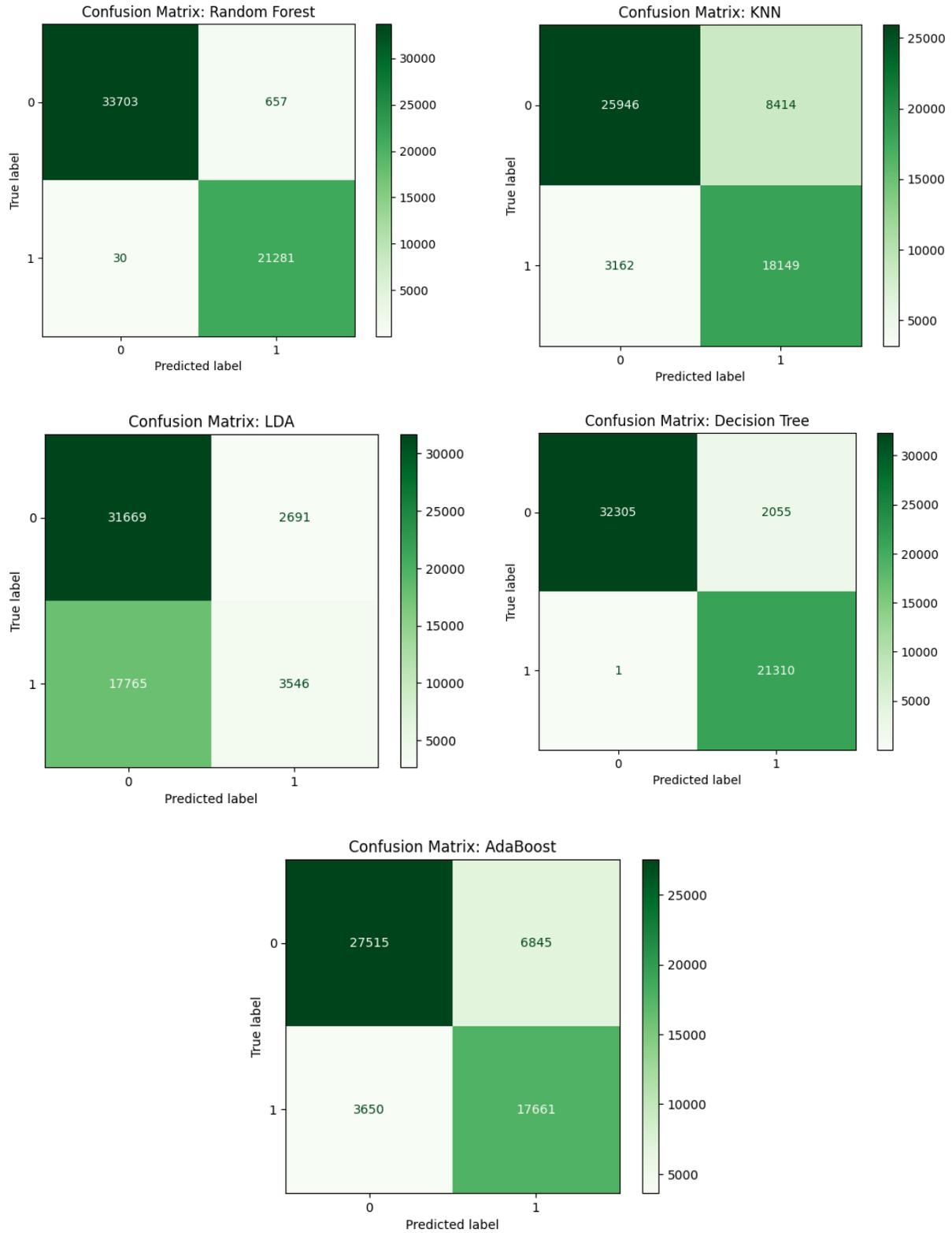

**Figure 5:** Confusion matrices for Random Forest, KNN, LDA, Decision Tree, and AdaBoost models, showing true positives, true negatives, false positives, and false negatives in Medicare fraud detection.

The confusion matrices in Figure 5 for Random Forest, KNN, LDA, Decision Tree, and AdaBoost models provide insights into the classification performance for Medicare fraud detection. Random Forest demonstrates exceptional performance, with 33,703 true negatives and 21,281 true positives. It achieves near-perfect recall, as only 30 fraudulent cases are misclassified as non-fraudulent. However, 657 non-fraudulent cases are incorrectly flagged as fraudulent, indicating a minor trade-off in precision. KNN shows moderate performance with 25,946 true negatives and 18,149 true positives. It misclassifies 3,162 fraudulent cases as non-fraudulent and incorrectly flags 8,414 non-fraudulent cases, leading to reduced recall and precision compared to Random Forest. LDA performs poorly, with a high number of false negatives (17,765 fraudulent cases classified as non-fraudulent) and a significant number of false positives (2,691 non-fraudulent cases flagged as fraudulent).

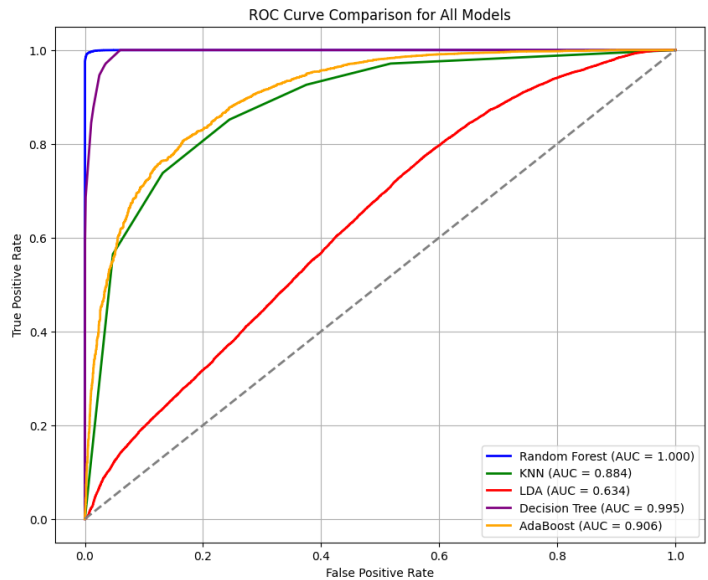

**Figure 6:** ROC curve comparison of Random Forest, KNN, LDA, Decision Tree, and AdaBoost models, highlighting their classification performance using AUC as the evaluation metric.

This indicates its inability to effectively separate fraudulent from non-fraudulent claims. Decision Tree performs similarly to Random Forest, with 32,305 true negatives and 21,310 true positives. It misclassifies only one fraudulent case as non-fraudulent, demonstrating perfect recall, though 2,055 false positives slightly affect precision. AdaBoost achieves balanced but moderate performance, with 27,515 true negatives and 17,661 true positives. However, 3,650 fraudulent cases and 6,845 non-fraudulent cases are misclassified, indicating a compromise between recall and precision.

The ROC curve compares the performance of the models in terms of their ability to distinguish between fraudulent and non-fraudulent Medicare claims. The area under the curve (AUC) is used as a performance metric, where a higher AUC indicates better discrimination ability. Random Forest achieves an AUC of 1.000, demonstrating perfect discrimination between classes. Its curve is tightly aligned with the top-left corner, indicating outstanding performance with minimal false positive and false negative rates. Decision Tree closely follows, with an AUC of 0.995. Its ROC curve is almost identical to Random Forest's, reflecting its strong capability to generalize while maintaining high sensitivity and specificity. AdaBoost achieves an AUC of 0.906, reflecting a reasonable balance between true positive and false positive rates. While it performs well overall, it lags behind Random Forest and Decision Tree, especially in the higher false positive rate range. KNN shows moderate performance with an AUC of 0.884. Its curve deviates more from the top-left corner, indicating a higher likelihood of false positives and negatives compared to the better-performing models. LDA performs poorly with an AUC of 0.634. Its ROC curve remains closest to the diagonal, signifying weak discrimination ability. This confirms LDA's limited utility in detecting fraudulent claims effectively.

## 5. Discussion

The detection of Medicare fraud remains a critical challenge for the healthcare sector due to the financial and social repercussions associated with fraudulent activities. This study underscores the potential of ML to address these challenges, offering a structured framework for effective fraud detection through advanced computational methods. However, several technical and practical considerations emerge from the analysis, which have implications for future research and implementation. One of the primary challenges in Medicare fraud detection is the extreme class imbalance in the dataset. Fraudulent claims represent a small fraction of the total claims, making it difficult for traditional models to effectively identify them without being biased toward the majority class. While resampling techniques such as SMOTE and hybrid methods like SMOTE-ENN can help alleviate this issue, they introduce complexities in ensuring that oversampling does not lead to overfitting. Moreover, the reliance on such techniques necessitates careful evaluation to balance model sensitivity and specificity. Another limitation lies in the complexity of Medicare claims data, which is highly structured and often includes hundreds of features ranging from demographic details to financial and procedural codes. Feature selection and dimensionality reduction play a crucial role in filtering out irrelevant data while retaining meaningful patterns indicative of fraud. However, the choice of methods for feature reduction, such as supervised selection or aggregation, significantly impacts the model's performance and interpretability. Additionally, the evolving nature of fraudulent schemes presents an ongoing challenge. Static models are prone to obsolescence as fraudsters adapt their strategies over time. The implementation of adaptive learning mechanisms, such as continuous retraining and updating with new data, is essential to maintain the efficacy of detection systems. However, this approach requires robust data pipelines and computational resources, which may not be feasible for all organizations.

Machine learning offers the advantage of scalability and adaptability, making it suitable for processing large Medicare datasets in real-time. By employing ensemble methods and deep learning architectures, detection systems can achieve improved accuracy and efficiency. However, the deployment of these models in real-world settings requires careful integration with existing fraud detection protocols and regulatory frameworks. For instance, high false-positive rates can lead to unnecessary investigations, straining healthcare providers and diverting resources from legitimate claims. Optimizing precision while maintaining high recall is essential for practical applications. Furthermore, ethical considerations play a significant role in the development and deployment of fraud detection systems. Transparency in model predictions, as well as mechanisms to address biases in data or algorithms, are critical to ensuring fairness. For instance, graph-based approaches, which analyze relationships between entities such as providers and beneficiaries, have shown promise in uncovering hidden patterns. However, these methods must be implemented with caution to avoid unjustly flagging legitimate providers due to inherent biases in the dataset.

The study highlights several avenues for future research. Hybrid models that combine the strengths of multiple machine learning techniques, such as deep learning and graph-based methods, could enhance detection capabilities by capturing complex interdependencies in Medicare claims data. Additionally, the integration of anomaly detection techniques with supervised learning models may further improve the system's ability to identify novel fraud patterns. Efforts should also focus on improving the interpretability of machine learning models. While advanced architectures like neural networks offer high accuracy, their "black box" nature limits their transparency. Explainable AI methods, such as feature attribution techniques, could bridge this gap, providing actionable insights to investigators while maintaining robust detection performance. Finally, collaboration between researchers, policymakers, and healthcare organizations is crucial to address the broader implications of Medicare fraud detection. By aligning technological advancements with practical needs and ethical considerations, machine learning can play a transformative role in safeguarding the integrity of healthcare systems.

## 6. Conclusion

This study demonstrates the application of ML techniques to effectively detect Medicare fraud, addressing challenges such as class imbalance, high-dimensional data, and evolving fraud patterns. Medicare fraud, which accounts for billions in financial losses annually, requires advanced computational solutions to identify deceptive practices such as upcoding, billing for unperformed services, and misrepresentation of diagnoses. By leveraging multiple ML models, this study achieved significant progress in identifying fraudulent claims while minimizing false positives. The Random Forest model emerged as the most effective, achieving a training accuracy of 99.2% and a validation accuracy of 98.8%. Its near-perfect recall of 99.9% indicates exceptional sensitivity to detecting fraudulent claims, while an F1-score of 98.4% underscores its balanced performance across all metrics. Similarly, the Decision Tree model achieved a validation accuracy of 96.3% and a recall of 100%, making it a competitive alternative, albeit with a slightly higher false-positive rate, as evidenced by a validation precision of 91.2%. Other models, such as KNN and AdaBoost, demonstrated moderate effectiveness. KNN achieved a validation accuracy of 79.2%, with a recall of 85.2% and an F1-score of 75.8%, indicating its tendency to produce more false positives. AdaBoost performed better, with a validation accuracy of 81.1% and an F1-score of 77.1%, showing its ability to balance recall and precision to some extent. However, these models lagged behind Random Forest and Decision Tree in overall performance. In contrast, LDA struggled significantly with a validation accuracy of only 63.3% and a recall of 16.6%, highlighting its limitations in handling the high-dimensional, imbalanced nature of the dataset. Its poor F1-score of 25.7% further emphasizes its inability to reliably detect fraudulent claims.

The findings highlight the importance of addressing key challenges in Medicare fraud detection. For example, balancing techniques like SMOTE and SMOTE-ENN effectively mitigated the issue of class imbalance, enabling models to focus on minority classes (fraudulent claims). Feature selection and dimensionality reduction were crucial in streamlining the dataset, preserving essential patterns while reducing complexity. Additionally, adaptive learning mechanisms ensured that models remained effective against evolving fraud patterns. From a business perspective, the framework introduces transformative advancements by significantly reducing financial losses through improved fraud detection accuracy and operational efficiency. By addressing class imbalance, leveraging advanced feature selection, and incorporating adaptive learning mechanisms, this approach ensures real-time fraud detection while reducing false positives and unnecessary investigations. The system's scalability and explainability foster trust among stakeholders, offering a proactive approach to fraud management that enhances compliance and resource allocation. Furthermore, the methodology positions organizations to gain a competitive edge by integrating robust fraud detection models into their operations, ensuring resilience and efficiency in protecting healthcare resources. The findings demonstrate the potential of this framework to improve decision-making, reduce costs, and safeguard the integrity of healthcare systems.

From a theoretical perspective, this study contributes significantly by addressing key challenges in Medicare fraud detection using machine learning techniques. It provides a structured framework that tackles extreme class imbalance through validated resampling methods like SMOTE, enhancing model sensitivity to minority class patterns. By implementing feature selection and dimensionality reduction techniques, it advances the understanding of high-dimensional data processing, preserving essential fraud indicators while minimizing computational complexity. Furthermore, the integration of adaptive learning mechanisms ensures that the models evolve alongside fraudulent schemes, maintaining robustness and effectiveness over time. The comparative analysis of ML algorithms, highlighting Random Forest and Decision Tree as top performers, establishes a theoretical basis for selecting models that balance accuracy, precision, and scalability. This study also emphasizes the importance of hybrid models and explainable AI to improve interpretability and adaptiveness in future fraud detection research.  Random Forest and Decision Tree proved to be the most reliable models for Medicare fraud detection, combining high accuracy, precision, and recall. Their scalability and robustness make them practical for real-world applications, where timely and accurate fraud detection is critical to safeguarding healthcare systems and resources. Future work should focus on integrating explainable AI techniques and hybrid models to further enhance detection performance and interpretability, bridging the gap between theoretical advancements and practical implementations in combating Medicare fraud.

**Future Work**

While this study demonstrates the effectiveness of machine learning models for Medicare fraud detection, several avenues for future research remain. One critical area is the integration of explainable AI (XAI) techniques, which would improve the interpretability of complex models like Random Forest and neural networks. Enhancing transparency in decision-making would ensure trust and enable investigators to better understand the reasoning behind flagged claims. Additionally, hybrid models that combine the strengths of different approaches, such as graph-based methods and deep learning architectures, could capture complex relationships in Medicare datasets more effectively. Another promising direction is the incorporation of real-time anomaly detection systems that leverage streaming data, enabling immediate identification of fraudulent activities. This would require robust pipelines for real-time data ingestion and processing. Expanding the application of this framework to include unsupervised learning techniques could help identify new, previously unclassified fraud patterns, further strengthening detection capabilities.

**Data is available**

Data is available and can be provided over the emails querying directly to the author at the corresponding author.

**Conflict of interest**

In this paper, the authors did not receive funding from any institution or company and declared that they do not have any conflict of interest.